\documentclass{article}

\usepackage{arxiv}

\usepackage[utf8]{inputenc} % allow utf-8 input
\usepackage[T1]{fontenc}    % use 8-bit T1 fonts
\usepackage{hyperref}       % hyperlinks
\usepackage{url}            % simple URL typesetting
\usepackage{booktabs}       % professional-quality tables
\usepackage{amsfonts}       % blackboard math symbols
\usepackage{nicefrac}       % compact symbols for 1/2, etc.
\usepackage{microtype}      % microtypography
\usepackage{lipsum}		% Can be removed after putting your text content
\usepackage{graphicx}
\usepackage{natbib}
\usepackage{doi}
\usepackage{algorithmic}
\usepackage{algorithm}
\usepackage{amsmath,amssymb,amsfonts}
\usepackage{multicol}
\usepackage{multirow}

\title{A Voting-Stacking Ensemble of Inception Networks for Cervical Cytology Classification}

\author{ \href{https://orcid.org/0009-0006-0823-5367}{\hspace{1mm}Linyi Qian} \\
	School of Computer and Information\\
	Hohai University\\
	Nanjing, China 211100 \\
	\texttt{qianlinyi@hhu.edu.cn} \\
	\And
	\href{https://orcid.org/0000-0001-5625-0402}{\hspace{1mm}Qian Huang} \\
	School of Computer and Information\\
	Hohai University\\
	Nanjing, China 211100 \\
	\texttt{huangqian@hhu.edu.cn} \\
}

\renewcommand{\shorttitle}{A Voting-Stacking Ensemble of Inception Networks for Cervical Cytology Classification}

\hypersetup{
pdftitle={A Voting-Stacking Ensemble of Inception Networks for Cervical Cytology Classification},
pdfauthor={Linyi Qian, Qian Huang},
pdfkeywords={Cervical cytology classification, Ensemble learning, Transfer learning, Stacking ensemble, Voting strategy},
}

\begin{document}
\maketitle

\begin{abstract}
	Cervical cancer is one of the most severe diseases threatening women's health. Early detection and diagnosis can significantly reduce cancer risk, in which cervical cytology classification is indispensable. Researchers have recently designed many networks for automated cervical cancer diagnosis, but the limited accuracy and bulky size of these individual models cannot meet practical application needs. To address this issue, we propose a Voting-Stacking ensemble strategy, which employs three Inception networks as base learners and integrates their outputs through a voting ensemble. The samples misclassified by the ensemble model generate a new training set on which a linear classification model is trained as the meta-learner and performs the final predictions. In addition, a multi-level Stacking ensemble framework is designed to improve performance further. The method is evaluated on the SIPakMed, Herlev, and Mendeley datasets, achieving accuracies of 100\%, 100\%, and 100\%, respectively. The experimental results outperform the current state-of-the-art (SOTA) methods, demonstrating its potential for reducing screening workload and helping pathologists detect cervical cancer.
\end{abstract}

% keywords can be removed
\keywords{Cervical cytology classification \and Ensemble learning \and Transfer learning \and Stacking ensemble}

\section{Introduction}
Cervical cancer is the fourth most frequently diagnosed cancer in women \citep{b1} and accounts for 4\% of all cancers diagnosed worldwide \citep{b2}. As a global health concern, the prevention and treatment of cervical cancer have been a hot topic in the medical community. With the emergence of HPV vaccines and screening techniques, the incidence of cervical cancer has dropped by more than half from the mid-1970s to the mid-2000s \citep{b2}. Nevertheless, It continues to be the second leading cause of cancer death in women aged 20 to 39 \citep{b3}. In addition, there has been no significant decrease in cervical cancer cases in low and middle-income countries (LMICs) \citep{b4}. On the one hand, the scarcity of HPV vaccines and the high cost make it unavailable for women in these regions. On the other hand, outdated screening techniques result in miss diagnosis, an area for which computer scholars can strive.

Currently, the Pap test is the most commonly used cervical cancer screening technique, which helps pathologists detect pre-cancerous cells and cervical lesions for early diagnosis and treatment. However, the Pap test relies on specialists to manually classify each cell on the slide, which can be time-consuming and labor-intensive. Therefore, computer-aided detection has become a promising alternative.

In the past decade, convolutional neural networks (CNNs) have made remarkable progress and breakthroughs, such as VGG \citep{b5}, Inception \citep{b6}, ResNet \citep{b7}, ResNeXt \citep{b8} and SENet \citep{b9}, especially in image classification. As a result, many researchers have applied them to cervical cell classification and achieved satisfactory results. \citep{b10} designed a CNN called DeepPap specifically for binary classification of cervical cells, reaching an accuracy of 98.3\% when evaluated on both the Pap smear and the liquid-based cytology (LBC) datasets. \citep{b11} presented deep learning classification methods applied to the SIPaKMeD dataset to establish a reference point for assessing forthcoming classification techniques and achieved the highest accuracy of 94.89\% using the resnet152 architecture. \citep{b12} combined InceptionV3 and artificial features, which effectively improves the accuracy of cervical cell recognition. Moreover, \citep{b13} have researched graph convolution networks (GCNs), which explored the potential relationships between cervical cell images by constructing graphs. They used GCN to enhance the discriminative ability of CNN features and achieved accuracies of 98.37\% on the SIPaKMeD dataset and 94.93\% on the private Motic dataset. Although these single-model methods have achieved good performance, there is still room for improvement. In recent years, many scholars have tried ensemble learning methods (detailed information provided in Section \ref{sec:related work}). They aggregated features extracted by different classifiers using a series of functions and applied them to the final prediction, thus further improving the classification accuracy.

This paper proposes a Voting-Stacking ensemble method for cervical cell classification. First of all, considering the size of the dataset, we uniformly resize the images in the dataset and use a combination of online and offline image augmentation. These images are fed into three Inception family models (each pre-trained on the ImageNet dataset), which serve as base learners. Then a voting strategy is applied to aggregate the outputs of these base learners. If there are contradictory predictions (e.g., misclassified samples), we send such samples to the meta-learner for further training and to make the final prediction. The proposed method is extensively evaluated on three public datasets: SIPaKMeD, Herlev, and Mendeley. Experimental results demonstrate that it exhibits good robustness and achieves the highest accuracy among all existing methods.

The main contributions of this paper are as follows:
\begin{itemize}
\item We design an ensemble of three homogeneous CNN models as base learners with size and number of parameters comparable to a deep model, but it can learn image features more effectively.

\item We propose a novel improved Stacking ensemble strategy called Voting-Stacking for cervical cell classification, which is the first time apply it in this area. We select contradictory samples after the ensemble and feed them to a meta-learner for retraining, thus coupling different features learned by base learners and improving classification accuracy.

\item We devise a multi-level Stacking ensemble framework based on the Voting-Stacking ensemble, which further improves the accuracy and provides a new direction for model ensemble in cervical cytology classification.

\item The proposed ensemble strategy is evaluated on three public cervical cell datasets using a range of metrics, and the results demonstrate that it outperforms state-of-the-art (SOTA) methods.
\end{itemize}

\section{Related work}
\label{sec:related work}
\subsection{Transfer Learning}
Transfer learning aims to improve the performance of target learners on target domains by transferring the knowledge contained in different but related source domains \citep{b14}. In this way, the dependence on many target-domain data can be reduced for constructing target learners \citep{b15}. Currently, the biggest challenge in medical image processing is the scarcity of publicly available datasets and the small size of these datasets, which results in the inability of models to learn the features of medical images thoroughly. To address the issue, many researchers have applied transfer learning in this field, using pre-trained models to conduct experiments.

The ensemble experiments on publicly available datasets mainly revolve around two datasets: SIPaKMeD and Herlev. \citep{b16} presented a study of transfer learning frameworks InceptionResNetV2, VGG19, DenseNet201, and Xception networks pre-trained on ImageNet, to classify cervical images using the SIPaKMeD dataset. The models achieved accuracies of 95.58\%, 94.91\%, 93.31\%, and 95.79\%, respectively. Similar to the former, \citep{b17} proposed an Internet of health things (IoHT)-driven deep learning framework for detecting cervical cancer in Pap smear images using transfer learning. They used pre-trained models like InceptionV3, VGG19, SqueezeNet, and ResNet50 to extract features for the binary classification of cervical cells and achieved an accuracy of 97.87\% on the Herlev dataset. In addition, the parameters and inference speed of the model are also important research directions in transfer learning. From this perspective, \citep{b18} designed a transfer learning-based deep EfficientNet model which is lightweight and dramatically reduces inference time. They also used the Herlev dataset to evaluate the method and achieved an accuracy of 99\%. Due to the limited number of publicly available datasets, some researchers have also conducted experiments on private datasets. \citep{b19} designed an adaptive pruning deep transfer learning model (PsiNet-TAP) for Pap smear image classification. They tested it on a private Pap smear dataset which contains 389 cervical cell images, and achieved more than 98\% accuracy. \citep{b20} used the pre-trained SqueezeNet architecture to handle the classification task and tested it on a private dataset that contains Pap smear images collected from the hospital, achieving an accuracy of 98.41\%. The previous discussions focus on the classification of single-cell. In the meantime, transfer learning for overlapping cell classification has also progressed. \citep{b21} investigated the use of transfer learning for the classification of overlapping cells and designed a transfer learning technique with the Alexnet framework implemented to improve classification accuracy. They finally achieved an accuracy of 99.86\% on the Cervix 93 dataset and an accuracy of 98.37\% on the Herlev dataset.

\subsection{Ensemble Learning}
Although transfer learning can further improve the accuracy, an individual model is limited by its architecture, and there is always an upper bound (i.e., Bayes error) which makes it increasingly difficult to improve the performance currently. Ensemble learning is the appropriate solution to the problem, combining multiple models to achieve better predictive performance by taking advantage of the strengths of each model and compensating for their weaknesses \citep{b22}.

Many scholars have recently applied ensemble learning to cervical cell classification. \citep{b23} researched the ensemble of machine learning models, proposing an ensemble method that combines three classifiers: decision tree (DT), nearest centroid (NC), and k-nearest neighbors (KNN), which are evaluated against the ISBI'14 Overlapping Cervical Cytology Image Segmentation Challenge dataset and achieved a precision of 98.8\%. \citep{b24} found that the CNN-based segmentation followed by an ensemble of some machine learning models achieved the best performance. They proposed an efficient automated hybrid framework for enhancing the cell classification accuracy of cervical cytology images. The model was evaluated on the Herlev dataset and achieved an average accuracy of 99.6\% for 2-class classification and 90.9\% for 4-class classification, respectively. 

In addition, research on the ensemble of CNN models is also a hot topic. \citep{b25} presented a simple and effective method for boosting the performance of trained CNNs by combining the scores (using the sum rule) of multiple CNNs into an ensemble. They achieved an accuracy of 94.88\% on the Herlev dataset. \citep{b26} set out to extract cells and cell clusters and classified those samples based on the Bethesda System for reporting cervical cytology, achieving an accuracy of 90.4\% and 91.6\% for the ensemble learning and deep learning methods when evaluated with a 5-fold cross-validation. Distinguishing from simple model stacking, utilizing mathematical methods for the ensemble can also improve classification accuracy. \citep{b27} proposed an ensemble scheme that used a fuzzy rank-based fusion of classifiers by considering two non-linear functions on the decision scores generated by base learners. The proposed framework achieved an accuracy of 95.43\% on the SIPaKMeD dataset and 95.43\% on the Mendeley LBC dataset. \citep{b28} proposed a novel ensemble method to minimize error values between the observed and the ground truth, which considers distances from the ideal solution in three different spaces. They achieved accuracy scores of 96.47\%, 98.58\%, and 99.62\% on the SIPaKMeD, Herlev, and Mendeley datasets.

Some scholars have also considered the combination of transfer learning and ensemble learning, which has led to the proposal of several high-performance models. \citep{b29} proposed an Ensembled Transfer Learning (ETL) framework to classify well, moderate, and poorly differentiated cervical histopathological images. They achieved an accuracy of 98.61\% on a private dataset labeled by two practical medical doctors. \citep{b30} proposed a transfer learning-based snapshot ensemble (TLSE) method by integrating snapshot ensemble learning with transfer learning in a unified and coordinated way. The method was evaluated on the Herlev dataset and achieved an accuracy of 65.65\% for the 7-class classification task. 

\section{Methods}
\begin{figure}[ht]
\centerline{\includegraphics[width=\columnwidth]{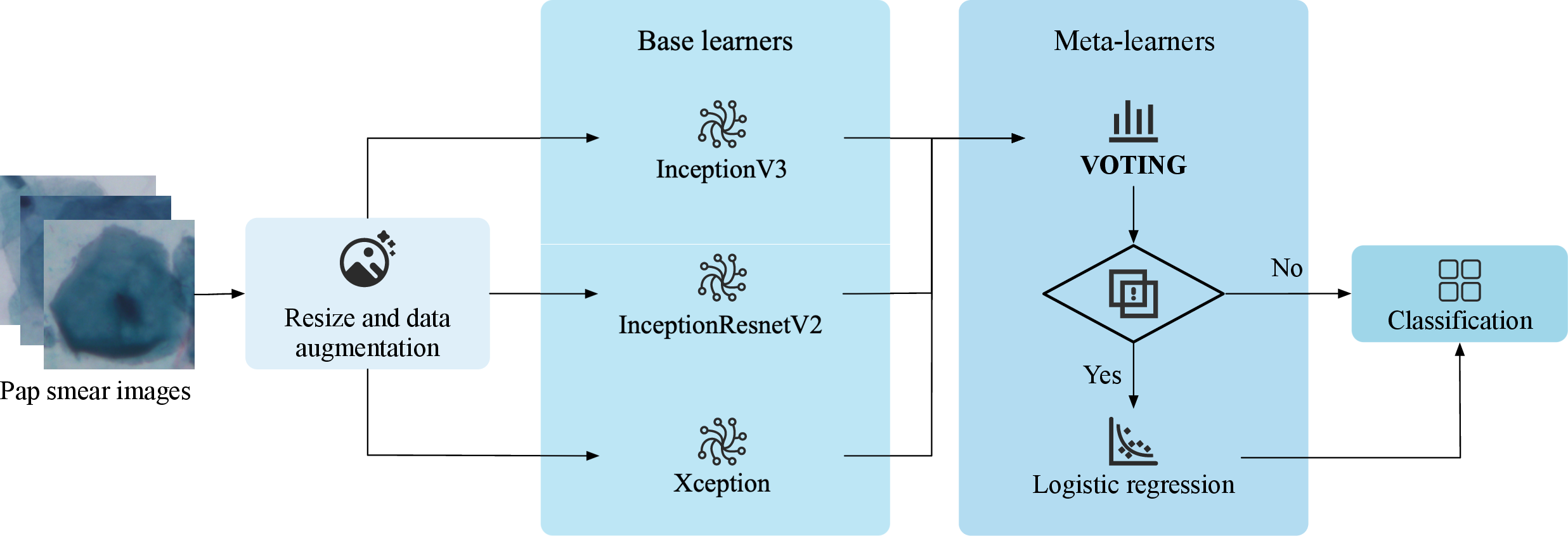}}
\caption{Overall workflow of the proposed method.}
\label{fig1}
\end{figure}
The proposed method consists of two stages. The first stage is data preprocessing, which involves resizing and data augmentation. The second stage mainly focuses on implementing the Voting-Stacking ensemble strategy. We utilize three Inception family CNNs, namely InceptionV3 \citep{b31}, InceptionResNetV2 \citep{b32}, and Xception \citep{b33}, to extract features from the input images and then aggregate the outputs of these models. For misclassified samples after the ensemble, a multinomial logistic regression model is used as the meta-learner to retrain and provide final predictions. The complete pipeline is illustrated in Fig. \ref{fig1}, and further details are described below.

\subsection{Data Preprocessing}
We find a common issue in processing public datasets: the sizes of images are inconsistent. Therefore, we perform uniform resizing of input images ($256\times256$) and modify the network input size to match the image size.

\citep{b34} proved CNN models becoming more competent in handling translation and rotation-related problems by experiments. In this paper, we employ offline and online augmentation methods to build a high-performing model. Offline data augmentation is used to expand the classes with a few samples, including vertical and horizontal flipping, which effectively triples the corresponding class size and achieves a relative balance in the datasets. Online augmentation techniques are also used, including random zooming, shifting, and rotation.

\subsection{Voting-Stacking Ensemble}
The Stacking ensemble strategy \citep{b35,b36} trains base learners on the initial training set and generates a new dataset for the meta-learner. In this new dataset, the output of the base learners is used as the sample input features, while the initial sample labels are still used as the sample labels. The pseudocode of the Stacking ensemble is shown in Algorithm \ref{alg1}. We propose improvements and optimizations to this method, and the specific implementation will be detailed in this section.

\begin{algorithm}
    \renewcommand{\algorithmicrequire}{\textbf{Input:}}
    \renewcommand{\algorithmicensure}{\textbf{Output:}}
    \caption{Pseudocode of Stacking Ensemble}
    \label{alg1}
    \begin{algorithmic}[1]
        \REQUIRE ~\\
        Training dataset $D=\lbrace (x_1,y_1),(x_2,y_2),\cdots,(x_m,y_m)\rbrace$; \\
        Base learners $M_1,M_2,\cdots,M_T$; \\
        Meta-learner $M$
        \FOR{$t=1,2,\cdots,T$}
        \STATE $h_t=M_t(D)$
        \ENDFOR
        \STATE $D^{\prime}=\varnothing$
        \FOR{$i=1,2,\cdots,m$}
        \FOR{$t=1,2,\cdots,T$}
        \STATE $z_{it}=h_t(x_i)$
        \ENDFOR
        \STATE $d_i=(z_{i1},z_{i2},\cdots,z_{iT})$
        \STATE $D^{\prime}=D^{\prime}\cup \lbrace(d_i,y_i)\rbrace$
        \ENDFOR
        \STATE $h^{\prime}=M(D^{\prime})$
        \ENSURE $H(x)=h^{\prime}(h_1(x),h_2(x),...,h_T(x))$
    \end{algorithmic}
\end{algorithm}

\subsubsection{Base Learners}
As the first step of the Stacking ensemble, the base learners need to learn the features of input medical images effectively, such as cell morphology, size, staining, nucleus-to-cytoplasm ratio, etc. Traditional machine learning methods rely on manual feature extraction, which requires massive human effort. Therefore, we choose deep learning models that can automatically extract features without additional preprocessing in this step.

The base learners in the stacking ensemble can be homogeneous or heterogeneous. When using heterogeneous classifiers, each classifier's weight must be considered carefully in experiments. Moreover, due to the vast number of existing neural networks, selecting networks with excellent ensemble performance takes much work. Therefore, we use homogeneous networks as base learners in this paper.

During experiments, we find that the Inception family networks show balanced performance across three datasets, even though the V3 version was proposed seven years ago. We infer that the different sizes of convolutional kernels in the Inception module can effectively learn the multi-scale features of images. Based on this, we choose three networks from the Inception family, InceptionV3 \citep{b31}, InceptionResNetV2 \citep{b32}, and Xception \citep{b33}, as base learners. Notably, the size and number of parameters of the ensemble are comparable to those of an individual deep network, but the performance of the former is better. The experimental results are provided in section \ref{sec:results} and the base learners are described as follows:

\paragraph{InceptionV3}
\citep{b31} proposed the third generation Inception-based CNN architecture, which improved the InceptionV2 in two aspects: firstly, optimizing the Inception module by introducing more convolutional kernel sizes (e.g., 8×8, 17×17, and 35×35) and using nested branches. Secondly, splitting the 2D convolution into two small 1D convolutions(e.g., a 7x7 convolution can be split into a 1x7 convolution and a 7x1 convolution). In other words, it uses the asymmetric convolution structure to process more spatial features and increase feature diversity while reducing computational complexity.
\paragraph{InceptionResNetV2}
\citep{b32} further researched the Inception network and proposed the InceptionResNetV2 architecture based on InceptionV3 by combining the Inception architecture with residual connections. This architecture integrates the advantages of two mainstream image recognition architectures: the Inception module can reduce the number of input channels, thereby reducing computational complexity, while the residual module can speed up network training and scale up the optimization process by fusing residual features with high-level features without concerning the vanishing gradient problem.
\paragraph{Xception}
\citep{b33} decoupled cross-channel and spatial correlation and proposed Xception by replacing Inception modules with depthwise separable convolutions (e.g., a depthwise convolution followed by a pointwise convolution). The architecture has the same parameters as InceptionV3 but performs better due to the more efficient use of model parameters.

\subsubsection{Meta-learners}
For the selection of the meta-learners, we opt for machine learning models. On the one hand, they have faster training speeds and fewer parameters than deep learning models. On the other hand, the necessary feature processing has already been performed by the CNNs in the base learning stage, allowing us to train the meta-learner directly without additional work.

The input attribute representation of the base learners and the meta-learning algorithm significantly impact the generalization performance of the Stacking ensemble. A study \citep{b39} has shown that using the output class probabilities of the base learners as the input attributes of the meta-learner and employing a multi-response linear regression (MLR) as the meta-learning algorithm yields better results. Based on this, we adopt multinomial logistic regression as the meta-learner, which is described as follows:

Given $n$ labeled samples $(\vec{x}_1,y_1),(\vec{x}_2,y_2),\cdots,(\vec{x}_n,y_n)$, where $\vec{x}_i=(x_1,x_2,\cdots,x_{M-1},b)$ is a feature vector of dimension $M$, with the last element denoted as $b$ representing the bias term. The label $y_i\in\lbrace1,2,\cdots,c\rbrace$ indicates the class, and each class has a corresponding weight vector $\vec{w}_i$. The probability of sample $x_i$ belonging to class $r$ can be calculated as follows:
\begin{equation}
P(r|\vec{x}_i)=\frac{\exp{(\vec{w}_r\cdot \vec{x}_i})}{\sum\limits_{j=1}^c\exp{(\vec{w}_j\cdot\vec{x}_i)}}
    \label{eq1}
\end{equation}
The probability distribution $D$ of the $i$-th sample and the loss function $L$ can be derived based on \eqref{eq1}:
\begin{equation}
    D=(P(1|\vec{x}_i),P(2|\vec{x}_i),\cdots,P(c|\vec{x}_i)),\sum\limits_{r=1}^cP(r|\vec{x}_i)=1
\end{equation}
\begin{equation}
    \begin{aligned}
L&=-\sum\limits_{i=1}^n\log{P(y_i|\vec{x}_i)}\\
&=\sum\limits_{i=1}^n\lbrace-\vec{w}_{y_i}\cdot\vec{x}_i+\log\sum_{j=1}^c\exp{(\vec{w}_j\cdot\vec{x}_i})\rbrace\\
&=\sum\limits_{i=1}^n\log(1+\sum\limits_{j\neq y_i}\exp{(\vec{w}_j\cdot\vec{x}_i-\vec{w}_{y_i}\cdot \vec{x}_{i})})
\end{aligned}
\end{equation}
Once the loss function is determined, the weights can be updated using gradient descent during training. The trained meta-learner will be used for the final predictions.

\subsubsection{Improved Stacking Ensemble}
The key innovation of this paper based on the Stacking ensemble strategy is the sift of inputs for the meta-learner, which only chooses contradictory samples. There are two advantages to doing this: First, it enhances the diversity of the data, as the meta-learner mainly focuses on samples that have wrong predictions made by base learners. In this case, retaining consistent samples can interfere with the model and affect final accuracy. Second, it significantly reduces the size of the meta-dataset, which can accelerate the training and inference speed of the meta-learner.

Assuming that the training set is $D=\lbrace(x_i,y_i)|_{i=1}^m\rbrace$, where $x_i$ represents the $i$-th image and $y_i\in\lbrace1,2,\cdots,c\rbrace$ represents the label of the corresponding image. The base learners are denoted as $M_1,M_2,\cdots,M_T$. First, each base learner is trained on the training set to obtain the corresponding classifier $h_t=M_t(D),t\in[1,2,\cdots,T]$. For the $i$-th image $x_i$, using the $t$-th classifier $h_t$ will generate a probability vector:
\begin{equation}
    P_{it}=(h^1_t(x_i),h_t^2(x_i),\cdots,h_t^c(x_i)),\sum\limits_{j=1}^ch^j_t(x_i)=1
\end{equation}
where  $h^j_t(x_i)$ represents the probability corresponding to the $j$-th class assigned by classifier $h_t$ for the $i$-th image. Then a soft voting strategy is applied to combine the output of each classifier and obtain the final predicted value:

\begin{equation}
\begin{aligned}
P_i&=\frac{1}{T}\sum\limits_{t=1}^TP_{it}=(H_1(x_i),H_2(x_i),\cdots,H_c(x_i))\\
&=\frac{1}{T}(\sum\limits_{t=1}^Th^1_t(x_i),\sum\limits_{t=1}^Th^2_t(x_i),\cdots,\sum\limits_{t=1}^Th^c_t(x_i)),\sum\limits_{j=1}^cH_{j}(x_i)=1
\end{aligned}
\end{equation}
\begin{equation}
    \hat{y}_i=\mathbf{argmin}(P_i)
\end{equation}
For the sample whose label does not match with the ground truth, we combine the probability vectors and corresponding label to form a training sample $(\mathcal{X}_i,y_i)$ for the meta-learner:
\begin{equation}
    (\mathcal{X}_i,y_i)=((P_{i1},P_{i2},\cdots,P_{it}),y_i),y_i\neq\hat{y}_i
\end{equation}
Iterating over the training set to generate a new training set:
\begin{equation}
    D^{\prime}=\lbrace (\mathcal{X}_i,y_i)|_{i=1}^{m^{\prime}})\rbrace
\end{equation}
The new training set $D^{\prime}$ will be used to train the meta-learner. Notably, we do not use $P_i$ to generate training samples because the meta-learner needs to learn the weight of each base learner rather than each class. The testing set can be processed in a similar way by combining the testing results of the base learners and feeding them to the meta-learner for final prediction (detailed information provided in Section \ref{sec:experiments}).

\subsubsection{Multi-level Stacking ensemble}
\begin{figure}[ht]
\centerline{\includegraphics[width=0.6\columnwidth]{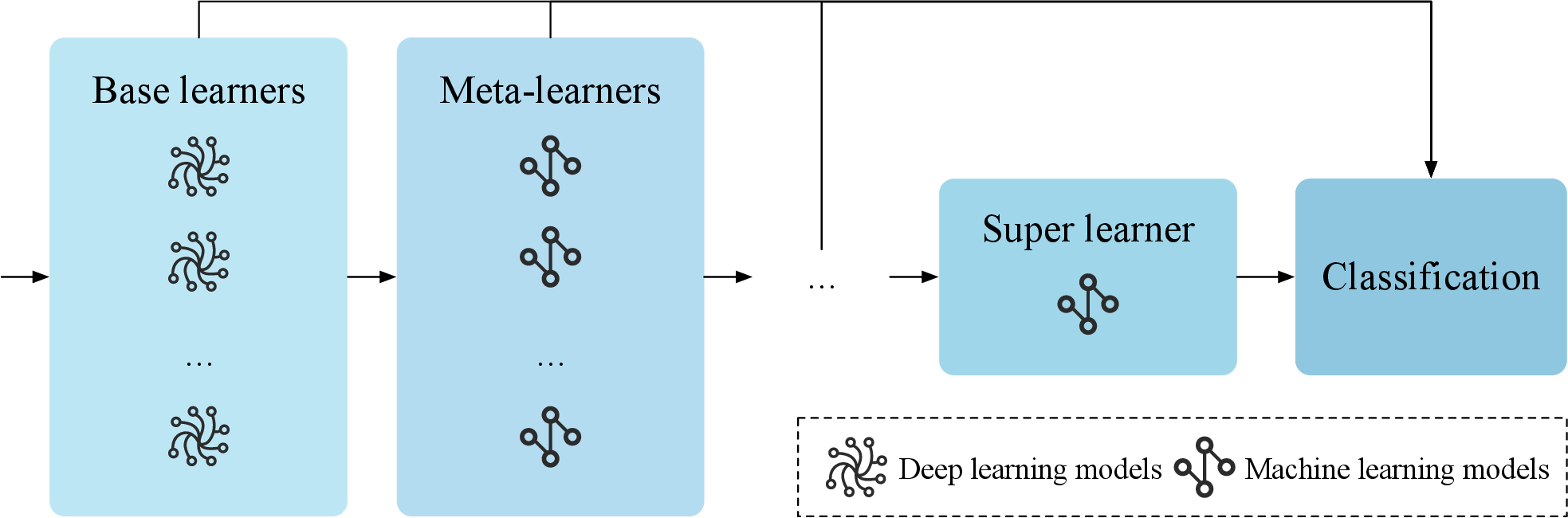}}
\caption{Framework of the multi-level Stacking ensemble.}
\label{fig10}
\end{figure}

When the dataset is relatively small, using only a two-level stacking strategy may not yield optimal classification results. Therefore, we propose a multi-level Stacking ensemble framework based on the improved Stacking ensemble. The whole framework is shown in Fig. \ref{fig10}. 

In the part of base learners (i.e., the first level of the ensemble), we choose deep learning models. Machine learning models are applied in the subsequent ensemble. During the experiments, we find that a three-level stacking ensemble is sufficient for most scenarios. As shown in Fig. \ref{fig10}, we use the output of base learners to train multiple meta-learners. Then, a super learner is applied to learn the output of these meta-learners and make the final predictions. The experimental results are provided in Section \ref{sec:results}.

\section{Experiments}
\label{sec:experiments}
\subsection{Datasets}
\begin{figure}[ht]
\centerline{\includegraphics[width=\columnwidth]{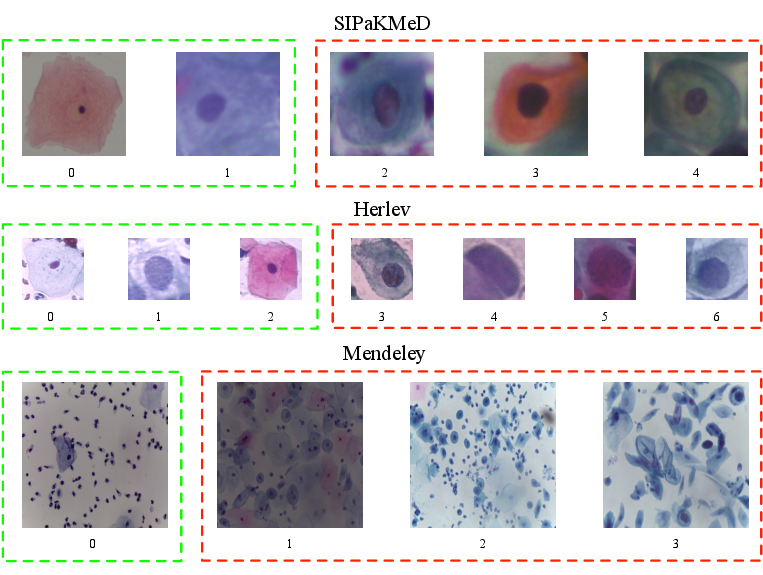}}
\caption{Examples of images from each class in the three public datasets, where the green dashed box represents normal cells, while the red dashed box represents abnormal cells.}
\label{fig4}
\end{figure}

In this paper, we evaluate the proposed method on three publicly available cervical cytology datasets:
\begin{enumerate}
    \item The SIPaKMeD Pap Smear dataset \citep{b40}
    \item The Herlev Pap Smear dataset \citep{b41}
    \item The Mendeley Liquid Based dataset \citep{b42}
\end{enumerate}
Detailed information about all datasets is listed in Table \ref{tab2} and examples of images are provided in Fig. \ref{fig4}.
\begin{table}[ht]
        \caption{Detailed description of three public datasets}
        \label{tab2}
        \resizebox{\linewidth}{!}{
            \begin{tabular}{ccccc}
                \toprule
                                          & Class & Index    & Cell type                                         & Number \\
                \midrule
                SIPaKMeD (total:4049)     & 0     & Normal   & Superficial-intermediate                          & 831    \\
                                          & 1     & Normal   & Parabasal                                         & 787    \\
                                          & 2     & Abnormal & Koilocytotic                                      & 825    \\
                                          & 3     & Abnormal & Dyskeratotic                                      & 813    \\
                                          & 4     & Abnormal & Metaplastic                                       & 793    \\
                \midrule
                Herlev (total:917)        & 0     & Normal   & Intermediate squamous epithelial                  & 70     \\
                                          & 1     & Normal   & Columnar epithelial                               & 98     \\
                                          & 2     & Normal   & Superficial squamous epithelial                   & 74     \\
                                          & 3     & Abnormal & Mild squamous non-keratinizing dysplasia          & 182    \\
                                          & 4     & Abnormal & Squamous cell carcinoma in-situ intermediate      & 150    \\
                                          & 5     & Abnormal & Moderate squamous non-keratinizing dysplasia      & 146    \\
                                          & 6     & Abnormal & Severe squamous non-keratinizing dysplasia        & 197    \\
                \midrule
                Mendeley LBC (total: 963) & 0     & Normal   & Negative for intraepithelial malignancy           & 613    \\
                                          & 1     & Abnormal & Low grade squamous intraepithelial lesion (LSIL)  & 163    \\
                                          & 2     & Abnormal & High grade squamous intraepithelial lesion (HSIL) & 113    \\
                                          & 3     & Abnormal & Squamous cell carcinoma (SCC)                     & 74     \\
                \bottomrule
            \end{tabular}
        }
\end{table}

\begin{figure}[t]
\centerline{\includegraphics[width=\columnwidth]{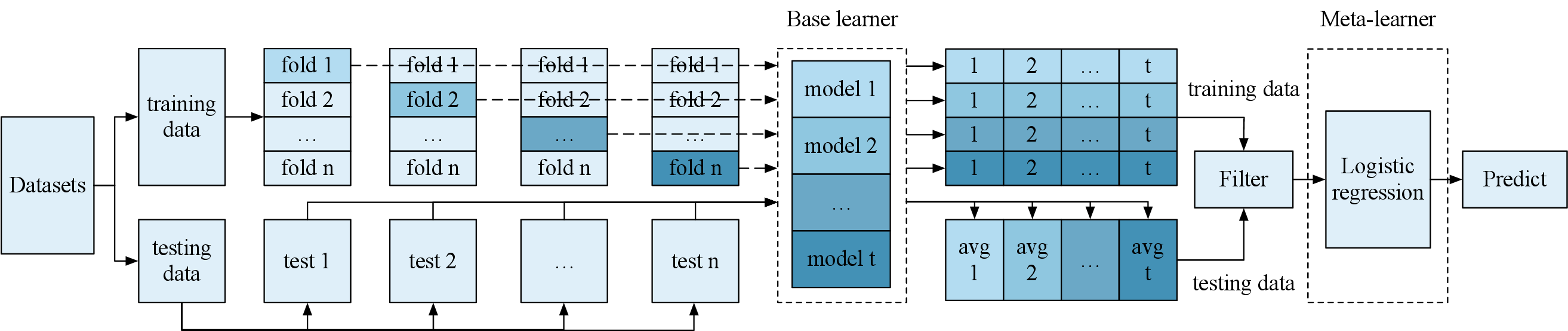}}
\caption{Pipeline of the evaluation stragety, where a specific color represents one fold.}
\label{fig5}
\end{figure}
\subsection{Evaluation Strategy}
\begin{figure}[ht]
\centerline{\includegraphics[width=0.6\columnwidth]{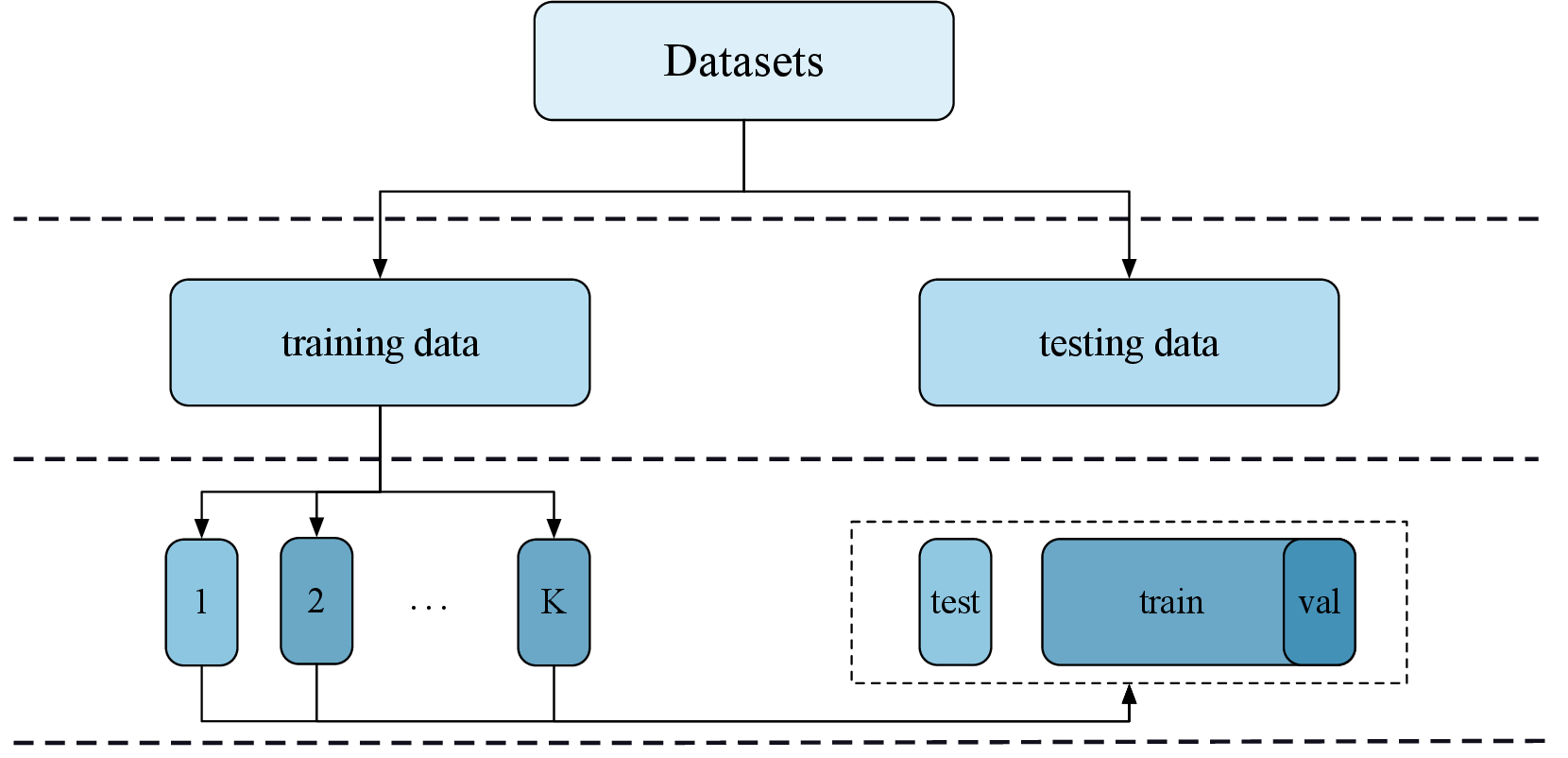}}
\caption{Dataset patitioning strategy.}
\label{fig3}
\end{figure}

In the training stage of base learners, we combine the Voting-Stacking strategy with $k$-fold cross-validation method and make optimizations for some details. The whole pipeline is provided in Fig. \ref{fig5} and the partitioning strategy is shown in Fig. \ref{fig3}.

First, the dataset is divided into a training set $D$ and a testing set $\tilde{D}$. When implementing $k$-fold cross-validation, the initial training set is divided into $k$ subsets of similar size, denoted as $D_1,D_2,\cdots,D_k$. Let $D_j$ and $\overline{D}_j=D\backslash D_j$ represent the testing set and the training set for the $j$-th fold, respectively. 

Consistent with the description in section III, we train $T$ base learners $M_1, M_2, \cdots, M_T$ on $\overline{D}_j$ and then test them on $D_j$. After filtering (e.g., choosing contradictory samples), the outputs are concatenated horizontally within one fold and vertically among $k$ folds to obtain the complete meta-training set. Here we focus on the processing of the testing set. All base learners trained in $k$-fold cross-validation are defined below:
\begin{equation}
h_{tj}=M_t(\overline{D}_j),t\in[1,2,\cdots,T],j\in[1,2,\cdots,k]
\end{equation}
For the $i$-th image $a_i$ in the testing set $\tilde{D}=\lbrace(a_i,b_i)|_{i=1}^{n}\rbrace$, the probability vector is the average of the outputs of $k$ base learners:

\begin{equation}
\begin{aligned}
P_{it}&=\frac{1}{k}\sum\limits_{j=1}^kP_{itj}\\
&=(\frac{1}{k}\sum\limits_{j=1}^kh_{tj}^1(x_i),\frac{1}{k}\sum\limits_{j=1}^kh_{tj}^2(x_i),\cdots,\frac{1}{k}\sum\limits_{j=1}^kh_{tj}^c(x_i))
\end{aligned}
\end{equation}
The remaining steps are the same as for the training set. Finally, we obtain the meta-training set $D^{\prime}=\lbrace (\mathcal{X}_i,y_i)|_{i=1}^{m^{\prime}})\rbrace$ and the meta-testing set $\tilde{D^{\prime}}=\lbrace(\mathcal{A}_{i},\beta_i)|_{i=1}^{n^{\prime}}\rbrace$.

Notably, we set a validation set in $k$-fold cross-validation to choose the model with the best performance for testing and apply the stratified sampling strategy to solve the the problem of imbalanced data distribution.

\subsection{Evaluation Metrics}
In this paper, we use four metrics to evaluate the performance of the method, namely accuracy, precision, recall, and F1 score. The definition of metrics is as follows:
\begin{equation}
    \mathbf{Accuracy}=\frac{\mathrm{TP}+\mathrm{TN}}{\mathrm{TP}+\mathrm{FP}+\mathrm{TN}+\mathrm{FN}}
\end{equation}
\begin{equation}
    \mathbf{Precision}=\frac{\mathrm{TP}}{\mathrm{TP}+\mathrm{FP}}
\end{equation}
\begin{equation}
    \mathbf{Recall}=\frac{\mathrm{TP}}{\mathrm{TP}+\mathrm{FN}}
\end{equation}
\begin{equation}
    \mathbf{F1}=\frac{2\times \mathbf{Precision}\times \mathbf{Recall}}{\mathbf{Precision}+ \mathbf{Recall}}
\end{equation}
where $\mathrm{TP}$ is the number of accurately labeled positive samples, $\mathrm{FP}$ represents the number of negative samples classified as positive, $\mathrm{TN}$ is the number of correctly classified negative samples, $\mathrm{FN}$ represents the number of positive instances predicted as negative.

\subsection{Experimental Configuration}

We follow the hyperparameter settings in \citep{b28} and make some optimizations. The learning rate, batch size, and loss function remain the same. For the number of epochs, they chose 70 because no validation set was set. However, in experiments, we find that the network overfitting occurs earlier, and since we include a validation set and save the model with the lowest validation loss during training, the number of epochs is reduced to 60. We introduce the AdamW optimizer with L2 regularization, which is computationally more efficient than Adam. In addition, we introduce the learning rate scheduler ReduceLROnPlateau, which detects the validation loss and automatically adjusts the learning rate when the validation loss does not decrease. As for the fold number, we choose 5 to make a balance between the time cost and the accuracy. The experimental configuration is shown in Table \ref{tab3} below:
\begin{table}[ht]
    \caption{The hyperparameters used for experiments}
\label{tab3}
\centering
        \begin{tabular}{cc}
            \toprule
            Hyperparameters         & Value/Method           \\
            \midrule
            Learning Rate           & 0.0001                 \\
            Batch Size              & 16                     \\
            Loss                    & Cross Entropy          \\
            Fold Number &5\\
            Epoch                   & 70$\rightarrow$\textbf{60}      \\
            Optimizer               & Adam$\rightarrow$\textbf{AdamW} \\
            Learning Rate Scheduler & \textbf{ReduceLROnPlateau}      \\
            \bottomrule
        \end{tabular}
\end{table}
\section{Results}
\label{sec:results}
\subsection{Model selection}
The selection of base learners and meta-learners in the Voting-Stacking ensemble significantly impacts the final accuracy. Therefore, we conduct a series of experiments focusing on model selection.

The comparative experiment results of base learners are shown in Table \ref{tab1}. We evaluate the performance of base learners and their ensembles based on three metrics: size, number of parameters, and average accuracy. Several popular CNN frameworks are compared, including the VGG, Inception, ResNet, EfficientNet, and ConvNeXt families. It can be observed that the Inception family models perform the best in accuracy. Besides, the Inception ensemble has a comparable size and number of parameters to an EfficientNetV2L or ConvNeXtBase model but achieves significantly higher accuracy. Therefore, we will use the three Inception models as our base learners in subsequent experiments.
\begin{table}[ht]
    \centering
    \caption{comparative experiment on base learners}
    \label{tab1}
    \centering
        \begin{tabular}{c|c|c|ccc}
            \hline
            \multirow{2}{*}{Model} & \multirow{2}{*}{Size(MB)} & \multirow{2}{*}{Parameters(M)} & \multicolumn{3}{c}{Accuracy(\%)}                                                     \\
            \cline{4-6}
                                   &                           &                                & SIPaKMeD                         & Herlev                  & Mendeley                \\
            \hline\hline
            Xception               & 88                        & 22.9                           & 96.99$\pm$0.66                   & 96.38$\pm$0.94          & 99.38$\pm$1.00          \\
            InceptionV3            & 92                        & 23.9                           & 94.86$\pm$0.89                   & 96.01$\pm$0.79          & 98.65$\pm$0.90          \\
            InceptionResNetV2      & 215                       & 55.9                           & 96.25$\pm$0.52                   & 94.78$\pm$2.30          & 99.59$\pm$0.60          \\
            Inception-Ensemble     & \textbf{395}              & \textbf{102.7}                 & \textbf{96.03$\pm$0.69}          & \textbf{95.72$\pm$1.34} & \textbf{99.21$\pm$0.83} \\
            \hline
            ResNet50               & 98                        & 25.6                           & 92.45$\pm$0.57                   & 91.54$\pm$0.38          & 96.12$\pm$0.52          \\
            ResNet101              & 171                       & 44.7                           & 93.24$\pm$0.72                   & 92.16$\pm$0.74          & 97.18$\pm$0.31          \\
            ResNet152              & 232                       & 60.4                           & 93.65$\pm$0.42                   & 92.86$\pm$0.16          & 97.33$\pm$0.75          \\
            ResNet-Ensemble        & 501                       & 130.7                          & 93.11$\pm$0.57                   & 92.19$\pm$0.43          & 96.88$\pm$0.53          \\
            \hline
            EfficientNetV2S        & 88                        & 21.6                           & 93.41$\pm$0.33                   & 92.62$\pm$0.24          & 98.26$\pm$0.43          \\
            EfficientNetV2M        & 220                       & 54.4                           & 94.32$\pm$1.02                   & 92.98$\pm$0.33          & 98.85$\pm$0.27          \\
            EfficientNetV2L        & 479                       & 119.0                          & 95.23$\pm$1.42                   & 93.12$\pm$0.56          & 99.12$\pm$0.64          \\
            EfficientNet-Ensemble  & 779                       & 195                            & 94.32$\pm$0.92                   & 92.91$\pm$0.38          & 98.74$\pm$0.45          \\
            \hline
            ConvNeXtSmall          & 192                       & 50.2                           & 94.32$\pm$0.15                   & 93.32$\pm$0.73          & 97.86$\pm$0.12          \\
            ConvNeXtBase           & 338                       & 88.5                           & 94.75$\pm$0.27                   & 94.42$\pm$0.54          & 98.85$\pm$0.42          \\
            ConvNeXtLarge          & 755                       & 197.7                          & 95.18$\pm$0.83                   & 94.86$\pm$0.42          & 99.45$\pm$0.33          \\
            ConvNeXt-Ensemble      & 1285                      & 342.7                          & 94.75$\pm$0.42                   & 94.20$\pm$0.56          & 98.72$\pm$0.29          \\
            \hline
            VGG13                 & 508                       & 133.1                          & 91.24$\pm$0.35                   & 90.35$\pm$0.32          & 95.54$\pm$0.76          \\
            VGG16                 & 528                       & 138.4                          & 92.28$\pm$1.22                   & 91.88$\pm$0.54          & 96.42$\pm$0.28          \\
            VGG19                  & 549                       & 143.7                          & 94.33$\pm$1.42                   & 92.09$\pm$0.94          & 96.88$\pm$0.58          \\
            VGG-Ensemble           & 1585                      & 415.2                          & 92.62$\pm$1.00                   & 91.44$\pm$0.60          & 96.28$\pm$0.54          \\
            \hline
        \end{tabular}
\end{table}

The comparative experiment results of meta-learners are shown in Table \ref{tab11}, where it can be observed that logistic regression performs the most balanced. It achieves the highest accuracy on both the SIPaKMeD and Mendeley datasets. In the multi-level stacking experiment, more meta-learners need to be introduced. Therefore, we choose the top three machine learning models with the highest accuracy: logistic regression, random forest classifier, and k-nearest neighbors classifier.

\begin{table}[ht]
    \caption{comparative experiment on meta-learners}
    \label{tab11}
    \centering
    \begin{tabular}{c|ccc}
        \hline
        \multirow{2}{*}{Model} & \multicolumn{3}{c}{Accuracy(\%)}                                                      \\
        \cline{2-4}            & SIPaKMeD                         & Herlev                  & Mendeley                 \\
        \hline\hline
        GaussianNB             & 98.66$\pm$0.17                   & 97.21$\pm$0.12          & 98.34$\pm$0.25           \\
        RidgeClassifier       & 98.93$\pm$0.22                   & 98.34$\pm$0.21          & 98.86$\pm$0.19           \\
        SVC                    & 99.01$\pm$0.24                   & 98.12$\pm$0.95          & 99.15$\pm$0.48           \\
        DecisionTreeClassifier & 99.10$\pm$0.36                   & 98.64$\pm$0.14          & 99.34$\pm$0.45           \\
        RandomForestClassifier & 99.18$\pm$0.32                   & 98.89$\pm$0.24          & 99.62$\pm$0.16           \\
        KNeighborsClassifier   & 99.30$\pm$0.21                   & \textbf{99.24$\pm$0.35} & 99.75$\pm$0.27           \\
        LogisticRegression    & \textbf{99.34$\pm$0.22}          & 99.12$\pm$0.07          & \textbf{100.00$\pm$0.00} \\
        \hline
    \end{tabular}
\end{table}
\subsection{Ablation study}
We conduct an ablation experiment to evaluate the importance of each component in the Voting-Stacking ensemble, and the results are shown in Table \ref{tab13}, from which several conclusions can be drawn: (1) More base learners means better classification accuracy. The model achieved the highest accuracy across three datasets when using three base learners. (2) Using only the meta-learner (e.g., general Stacking) performs better than using only the voting ensemble (e.g., general ensemble), which demonstrates that the Stacking ensemble is more efficient than traditional ensemble methods. (3) The Voting-Stacking ensemble improves the accuracy of the general Stacking ensemble, proving the superiority of our approach.
\begin{table}[ht]
    \centering
    \caption{ablation study for Voting-stacking ensemble}
    \label{tab13}
    \begin{tabular}{c|c|c|ccc}
        \hline
        \multirow{2}{*}{Number of base learners} & \multirow{2}{*}{Voting} & \multirow{2}{*}{Meta-learner} & \multicolumn{3}{c}{Accuracy(\%)}                     \\
        \cline{4-6}
                           &                                  &                               & SIPaKMeD                          & Herlev & Mendeley \\
        \hline\hline
        1                             & $\checkmark$                     & $\checkmark$                  & 97.89                             & 97.12  & 99.26    \\
        \hline
        2                             & $\checkmark$                     & $\checkmark$                  & 98.62                             & 98.24  & 99.45    \\
        \hline
        3                             & $\checkmark$                     &                               & 99.34                             & 99.08  & 99.75    \\
        \hline
        3                             &                                  & $\checkmark$                  & 99.51                             & 99.42  & 99.75    \\
        \hline
        3                             & $\checkmark$                     & $\checkmark$                  & \textbf{99.75}                             & \textbf{100.00} & \textbf{100.00}   \\
        \hline
    \end{tabular}
\end{table}
\subsection{Experimental results on SIPaKMeD}
Table \ref{tab4} presents the experimental results of the proposed method on the SIPaKMeD Pap Smear dataset. For the individual models (i.e., base learners), we calculate the mean and standard deviation of five tests due to adopting 5-fold cross-validation during training. However, our method is only tested once after the training of the meta-learner, hence there is only one value. Notably, the class distribution in the SIPaKMeD dataset is relatively balanced (as shown in Table \ref{tab2}, with around 800 images per class), so we do not use offline data augmentation on this dataset. The results show that the Voting-Stacking ensemble can improve performance by three percentage points compared to the three individual models. The performance of the ensemble model in all metrics is significantly better than that of individual models, demonstrating the superiority of our approach.
Additionally, to verify the robustness of the model, we evaluate it on three different sizes of testing sets (i.e., 10\%, 20\%, and 30\% of the dataset). The results show that the testing accuracy of the ensemble model is above 99\% for all testing sets, achieving 99.75\%, 99.51\%, and 99.34\%, respectively. When implementing the three-level Stacking ensemble, we further improve the accuracies to 100\%, 99.88\%, and 99.67\%, respectively.

\begin{figure}[htbp]
    \centering
    \begin{minipage}[c]{0.3\linewidth}
        \centering
        \includegraphics[width=\linewidth]{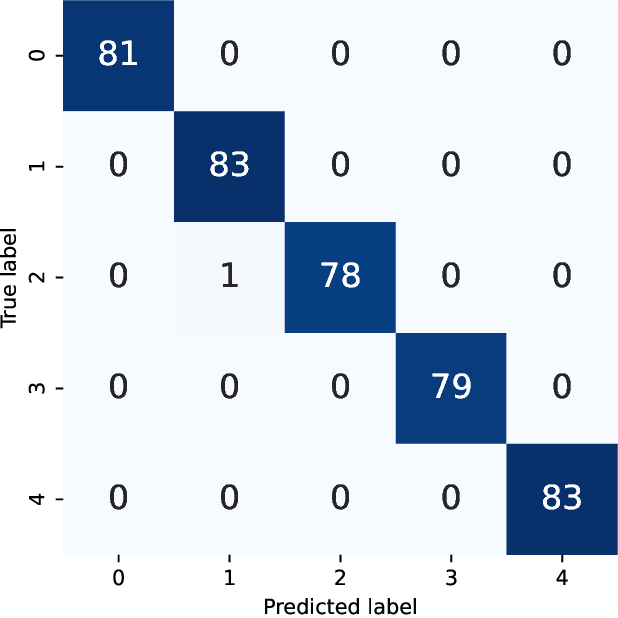}
    \end{minipage}
    \begin{minipage}[c]{0.3\linewidth}
        \centering
        \includegraphics[width=\linewidth]{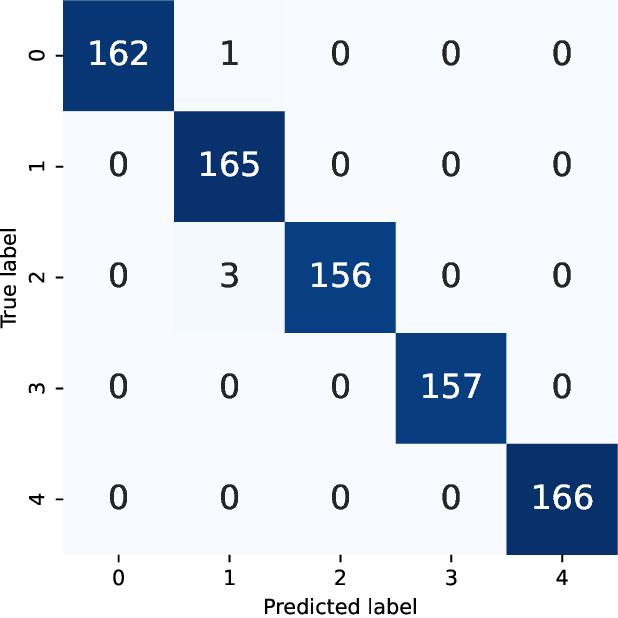}
    \end{minipage}
    \begin{minipage}[c]{0.3\linewidth}
        \centering
        \includegraphics[width=\linewidth]{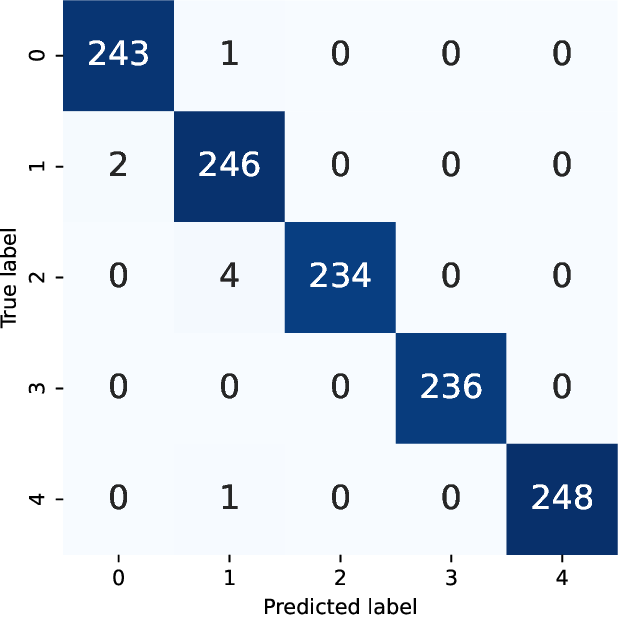}
    \end{minipage}
    \caption{Confusion matrices of results on SIPaKMeD, the index can be referred to the classes in Table \ref{tab2} and the sizes of the testing set from left to right are 10\%, 20\%, and 30\% of the dataset, respectively}
    \label{fig6}
\end{figure}

\begin{table}[ht]
        \caption{Results of 5-fold cross-validation on SIPaKMeD}
        \label{tab4}
        \centering
            \begin{tabular}{c|cccc}
                \hline
                \multirow{2}{*}{Method} & \multicolumn{4}{c}{5-class}                                                    \\
                \cline{2-5}
                                        & Accuracy(\%)                    & Precision(\%)      & Recall(\%)         & F1(\%)             \\
                \hline\hline
                InceptionV3(10\%)       & 94.86$\pm$0.89              & 94.95$\pm$0.85 & 94.92$\pm$0.87 & 94.85$\pm$0.90 \\
                InceptionV3(20\%)       & 93.88$\pm$0.59              & 94.02$\pm$0.58 & 93.97$\pm$0.57 & 93.84$\pm$0.60 \\
                InceptionV3(30\%)       & 92.82$\pm$0.69              & 92.92$\pm$0.64 & 92.90$\pm$0.68 & 92.84$\pm$0.70 \\
                InceptionResNetV2(10\%) & 96.25$\pm$0.52              & 96.32$\pm$0.49 & 96.30$\pm$0.51 & 96.24$\pm$0.52 \\
                InceptionResNetV2(20\%) & 96.04$\pm$0.27              & 96.15$\pm$0.26 & 96.11$\pm$0.26 & 96.04$\pm$0.27 \\
                InceptionResNetV2(30\%) & 94.65$\pm$0.86              & 94.73$\pm$0.80 & 94.71$\pm$0.84 & 94.67$\pm$0.85 \\
                Xception(10\%)          & 96.99$\pm$0.66              & 97.04$\pm$0.63 & 97.03$\pm$0.64 & 96.99$\pm$0.66 \\
                Xception(20\%)          & 96.69$\pm$0.71              & 96.75$\pm$0.70 & 96.74$\pm$0.71 & 96.69$\pm$0.71 \\
                Xception(30\%)          & 96.02$\pm$0.49              & 96.08$\pm$0.50 & 96.05$\pm$0.48 & 96.05$\pm$0.50 \\
                Ours(10\%)     & 99.75              & 99.75 & 99.76 & 99.75 \\
                Ours(20\%)     & 99.51                       & 99.53          & 99.50          & 99.51          \\
                Ours(30\%)     & 99.34                       & 99.34          & 99.36          & 99.35          \\
                Three-level Stacking(10\%)&\textbf{100.00}&\textbf{100.00}&\textbf{100.00}&\textbf{100.00}\\
                Three-level Stacking(20\%)&99.88&99.88&99.88&99.88\\
                Three-level Stacking(30\%)&99.67&99.67&99.68&99.67\\
                \hline
            \end{tabular}
\end{table}
In Fig. \ref{fig6}, we present the confusion matrices of the proposed method (two-level Stacking and without data augmentation) on three different testing sets. It can be observed that the parabasal cells (Class 1) are the most easily confused compared to other cells. The main reason for the decrease in accuracy is that the ensemble model misclassifies images of other classes as Class 1. We speculate that parabasal cells have similar morphological characteristics to other cells, making it difficult to distinguish them. The ensemble model performs best in distinguishing dyskeratotic cells (Class 3) as they have apparent pathological features.

Table \ref{tab5} compares our method with the state-of-the-art (SOTA) methods on the SIPaKMeD dataset. FuzzyRankEnsemble \citep{b27}, FuzzyDistanceEnsemble \citep{b28}, and PCAandGWO \citep{b44} are three ensemble learning methods; XCiT-S24 \citep{b45} and Swin-B \citep{b46} are pure individual models without any change, and the rest are individual models with some improvements. Whether compared with ensemble methods or individual networks, our method has achieved the best performance with accuracy, precision, recall, and F1 scores of 100\%, 100\%, 100\%, and 100\%, respectively.

\begin{table}[ht]
        \caption{Comparison with SOTA methods on SIPaKMeD}
        \label{tab5}
        \centering
            \begin{tabular}{ccccc}
                \hline
                Model                     & Accuracy(\%)        & Precision(\%)       & Recall(\%)          & F1(\%)              \\
                \hline\hline
                FuzzyRankEnsemble \citep{b27}       & 95.43           & 95.34           & 95.38           & 95.36           \\
                FuzzyDistanceEnsemble \citep{b28}   & 96.96           & 96.92           & 96.97           & 96.91           \\
                CytoBrain \citep{b43}                & 97.80           & 98.76           & 97.80           & 98.28           \\
                PCAandGWO \citep{b44}                & 97.87           & 98.56           & 99.12           & 98.89           \\
                XCiT-S24 \citep{b45}                  & 97.93           & 97.94           & 97.90           & 97.92           \\
                Swin-B \citep{b46}                   & 98.13           & 98.12           & 98.10           & 98.11           \\
                ExemplarPyramid \citep{b47} & 98.26           & 98.27           & 98.28           & 98.28           \\
                MTFFM \citep{b48}                     & 98.67           & 98.69           & 98.65           & 98.67           \\
                MLNet \citep{b49}                     & 99.31           & 99.29           & 99.26           & 99.24           \\
                Ours                      & \textbf{100.00} & \textbf{100.00} & \textbf{100.00} & \textbf{100.00} \\
                \hline
            \end{tabular}
\end{table}
\subsection{Experimental results on Herlev}
We conduct the same experiments on another Pap smear dataset called Herlev, which is first split into two classes: normal (index-0) and abnormal (index-1). The number of samples in the later class (675) is almost three times that of the former (242), making the initial distribution of the dataset highly imbalanced. To address this issue, we employ offline data augmentation on the normal class, increasing its size by three times. As shown in Table \ref{tab6}, the performance of the individual models on this dataset is worse than that in SIPaKMeD. However, the proposed method can still fill the gap by achieving 100\%, 99.46\%, and 99.64\% on three testing sets. Furthermore, with the offline data augmentation (denoted by *) and the three-level Stacking ensemble, the accuracy of the latter two increases to 100\% and 100\%, further confirming the effectiveness of our method.
\begin{table}[ht]
        \caption{Results of 5-fold cross-validation on Herlev}
        \label{tab6}
        \centering
            \begin{tabular}{c|cccc}
                \hline
                \multirow{2}{*}{Method} & \multicolumn{4}{c}{2-class}                                                       \\
                \cline{2-5}
                                        & Accuracy(\%)                    & Precision(\%)       & Recall(\%)          & F1(\%)              \\
                \hline\hline
                InceptionV3(10\%)       & 94.57$\pm$1.68              & 93.56$\pm$2.54  & 92.28$\pm$2.18  & 92.86$\pm$2.22  \\
                InceptionV3(20\%)       & 94.78$\pm$2.14              & 92.83$\pm$2.65  & 94.10$\pm$2.44  & 93.46$\pm$2.77  \\
                InceptionV3(30\%)       & 96.01$\pm$0.79              & 96.30$\pm$1.13  & 93.43$\pm$1.45  & 94.71$\pm$1.09  \\
                InceptionResNetV2(10\%) & 92.83$\pm$2.34              & 91.24$\pm$2.83  & 90.02$\pm$3.60  & 90.53$\pm$3.14  \\
                InceptionResNetV2(20\%) & 94.78$\pm$2.30              & 92.98$\pm$3.24  & 94.10$\pm$2.44  & 93.46$\pm$2.77  \\
                InceptionResNetV2(30\%) & 91.88$\pm$4.23              & 91.16$\pm$5.86  & 87.55$\pm$5.62  & 89.10$\pm$5.65  \\
                Xception(10\%)          & 93.26$\pm$2.11              & 92.83$\pm$2.57  & 89.51$\pm$3.60  & 90.88$\pm$3.01  \\
                Xception(20\%)          & 94.02$\pm$1.33              & 91.92$\pm$2.10  & 93.33$\pm$1.25  & 92.51$\pm$1.55  \\
                Xception(30\%)          & 96.38$\pm$0.94              & 96.13$\pm$1.32  & 94.64$\pm$2.14  & 95.25$\pm$1.33  \\
                Ours(10\%)              & \textbf{100.00}             & \textbf{100.00} & \textbf{100.00} & \textbf{100.00} \\
                Ours(20\%)              & 99.46                       & 99.00           & 99.63           & 99.31           \\
                                Ours(30\%)              & 99.07                       & 99.07           & 99.07           & 99.53           \\
                Ours*(20\%)             & \textbf{100.00}             & \textbf{100.00} & \textbf{100.00} & \textbf{100.00} \\
                Ours*(30\%)             & 99.76              & 99.77  & 99.75  & 99.76  \\
                                Three-level Stacking*(30\%)& \textbf{100.00} & \textbf{100.00} & \textbf{100.00} & \textbf{100.00} \\
                \hline
            \end{tabular}
\end{table}

\begin{figure}[htbp]
    \begin{minipage}[c]{0.3\linewidth}
        \centering
        \includegraphics[width=\linewidth]{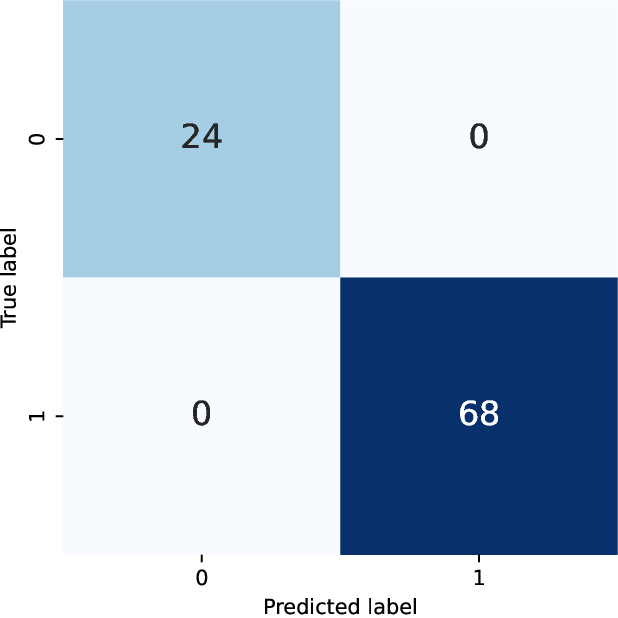}
    \end{minipage}
    \begin{minipage}[c]{0.3\linewidth}
        \centering
        \includegraphics[width=\linewidth]{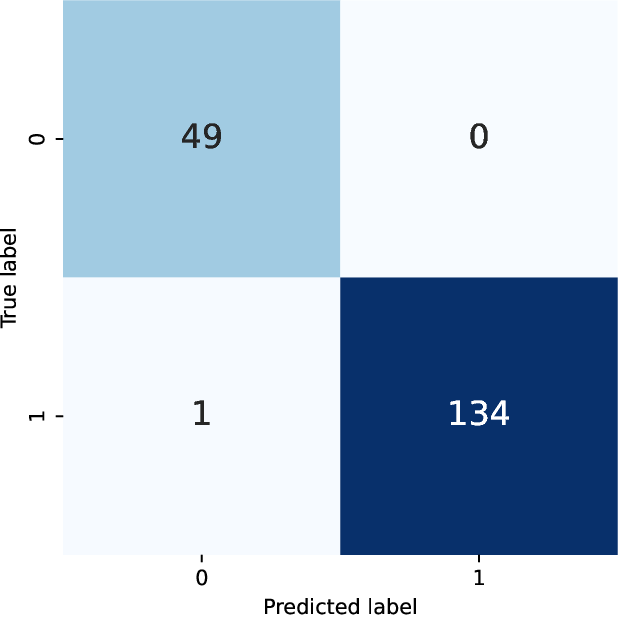}
    \end{minipage}
    \begin{minipage}[c]{0.3\linewidth}
        \centering
        \includegraphics[width=\linewidth]{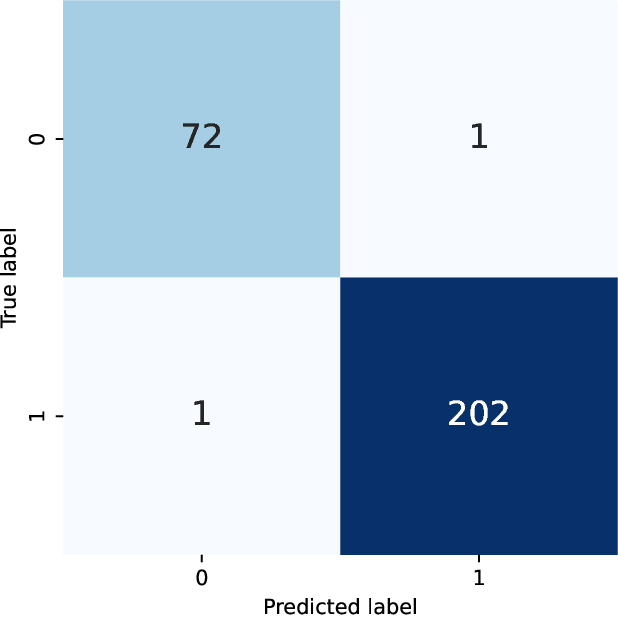}
    \end{minipage}
    \caption{Confusion matrices of results on Herlev, the index can be referred to the classes in Table \ref{tab2} and the sizes of the testing set from left to right are 10\%, 20\%, and 30\% of the dataset, respectively}
    \label{fig7}
\end{figure}

\begin{table}[ht]
        \caption{Comparison with SOTA methods on Herlev}
        \label{tab7}
        \centering
            \begin{tabular}{ccccc}
                \hline
                Model                   & Accuracy(\%)        & Precision(\%)       & Recall(\%)          & F1(\%)              \\
                \hline\hline
                DeepPap \citep{b10}                 & 98.30           & 99.40           & 98.20           & 98.80            \\
                PCAandGWO \citep{b44}              & 98.32           & 98.66           & 97.65           & 98.12           \\
                FuzzyDistanceEnsemble \citep{b28} & 98.58           & 98.65           & 98.53           & 98.58           \\
                DeepCervix \citep{b50}              & 98.91           & 99.50           & 98.00           & 98.50           \\
                MLNet \citep{b49}                   & 99.36           & 99.35           & 99.36           & 99.28           \\
                PSPM \citep{b24}                    & 99.70           & 99.20           & 99.80           & 99.30           \\
                Ours                    & \textbf{100.00} & \textbf{100.00} & \textbf{100.00} & \textbf{100.00} \\
                \hline
            \end{tabular}
\end{table}
Confusion matrices (two-level Stacking and without data augmentation) are provided in Fig. \ref{fig7}. It can be seen that the ensemble model has a solid discriminatory power for normal cells, and no errors are made on three testing sets. However, the ensemble model may make mistakes in identifying abnormal cells. This is due to the small number of training samples in the latter two scenarios, which means that the ensemble model does not learn enough features, resulting in misclassifying some abnormal cervical cells as normal cells. 

In Table \ref{tab7}, we compare our methods with existing cervical cancer classification techniques on the Herlev dataset. It can be found that all of these methods have achieved remarkable results on the 2-class classification task, and our method performs slightly better than PSPM, with all metrics reaching 100\%.

\subsection{Experimental results on Mendeley}
To further verify the generalization ability of the proposed method, we evaluate it on the Mendeley liquid-based dataset. This dataset is different from the first two, as each image contains multiple cells instead of a single cell, and the class of the image is determined by the features of the multi-cell. The results are shown in Table \ref{tab9}. Even without data augmentation, we can see that the proposed method achieves 100\% accuracy on all three testing sets. We speculate that multi-cell images contain more features, thereby enhancing the discriminatory power of the network. The comparison with the SOTA methods is shown in Table \ref{tab10}. 
\begin{table}[ht]
        \caption{Results of 5-fold cross-validation on Mendeley}
        \label{tab9}
        \centering
            \begin{tabular}{c|cccc}
                \hline
                \multirow{2}{*}{Method} & \multicolumn{4}{c}{4-class}                                                       \\
                \cline{2-5}
                                        & Accuracy(\%)                    & Precision(\%)       & Recall(\%)          & F1(\%)              \\
                \hline\hline
                InceptionV3(10\%)       & 95.88$\pm$1.46              & 93.49$\pm$4.40  & 88.13$\pm$3.78  & 89.49$\pm$4.35  \\
                InceptionV3(20\%)       & 98.65$\pm$0.90              & 97.37$\pm$1.71  & 96.50$\pm$2.70  & 96.84$\pm$2.30  \\
                InceptionV3(30\%)       & 95.22$\pm$0.77              & 90.24$\pm$2.65  & 90.23$\pm$0.53  & 89.39$\pm$0.98  \\
                InceptionResNetV2(10\%) & 97.32$\pm$0.51              & 97.17$\pm$0.51  & 91.88$\pm$1.53  & 93.62$\pm$1.31  \\
                InceptionResNetV2(20\%) & 99.38$\pm$1.00              & 99.20$\pm$1.27  & 98.00$\pm$3.23  & 98.40$\pm$2.62  \\
                InceptionResNetV2(30\%) & 96.33$\pm$0.94              & 92.06$\pm$2.47  & 92.34$\pm$1.54  & 91.71$\pm$1.87  \\
                Xception(10\%)          & 97.53$\pm$0.82              & 97.09$\pm$0.81  & 92.50$\pm$2.50  & 93.95$\pm$2.07  \\
                Xception(20\%)          & 99.59$\pm$0.60              & 99.21$\pm$1.25  & 98.96$\pm$1.41  & 99.07$\pm$1.33  \\
                Xception(30\%)          & 95.50$\pm$0.61              & 90.02$\pm$0.96  & 92.32$\pm$0.66  & 90.28$\pm$1.15  \\
                Ours(10\%)              & \textbf{100.00}             & \textbf{100.00} & \textbf{100.00} & \textbf{100.00} \\
                Ours(20\%)              & \textbf{100.00}             & \textbf{100.00} & \textbf{100.00} & \textbf{100.00} \\
                Ours(30\%)              & \textbf{100.00}             & \textbf{100.00} & \textbf{100.00} & \textbf{100.00} \\
                \hline
            \end{tabular}
\end{table}

\begin{table}[ht]
        \caption{Comparison with SOTA methods on Mendeley}
        \label{tab10}
        \centering
            \begin{tabular}{ccccc}
                \hline
                Model                     & Accuracy(\%)        & Precision(\%)       & Recall(\%)          & F1(\%)              \\
                \hline\hline
                c-CNN \citep{b51}                     & 96.89           & 93.38           & 93.75           & 94.15           \\
                FuzzyDistanceEnsemble \citep{b28}   & 99.23           & 99.13           & 99.23           & 99.18           \\
                MLNet \citep{b49}                     & 99.36           & 99.35           & 99.32           & 99.32           \\
                ExemplarPyramid \citep{b47} & 99.47           & 99.26           & 98.21           & 98.73           \\
                PCAandGWO \citep{b44}                & 99.47           & 99.14           & 99.27           & 99.20           \\
                FuzzyDistanceEnsemble \citep{b28}   & 99.68           & 99.34           & 99.87           & 99.60           \\
                Ours                      & \textbf{100.00} & \textbf{100.00} & \textbf{100.00} & \textbf{100.00} \\
                \hline
            \end{tabular}
\end{table}
\section{Discussion}
In contrast to recent ensemble methods, the method proposed in this paper provides detailed explanations for each step of ensemble building (including data preprocessing, model selection, ensemble optimization, evaluation strategies, etc.), and the above experimental results comprehensively verify the effectiveness of our method.

The Voting-Stacking ensemble optimizes the general Stacking ensemble strategy. When integrating base learners, only mispredicted samples are selected and sent to the meta-learner for training. In the meta-learning stage, weights are re-assigned for each class, and the weight interference of correctly predicted samples is eliminated, further improving accuracy. After in-depth analysis, future work could be improved as follows:
% However, ensemble learning also has some drawbacks. Firstly, the base learners require a longer training time than an individual model. Secondly, the ensemble model inevitably leads to an increase in the number of parameters, occupying more memory and graphics memory during training and making higher demands on experimental configurations, thereby increasing research costs.

\begin{itemize}
    \item For experimental convenience, we choose homogeneous learners, which may have weaker recognition ability for specific features in the classification process, such as the parabasal class in SIPaKMeD and the abnormal class in Herlev. Therefore, if we can select heterogeneous learners with strong discriminative power for these two types of cervical cells and assign appropriate weights, the performance of the ensemble model would be better.
    \item In the Voting-Stacking ensemble process, a deep learning model is used in the meta-learner training stage, while only machine learning models are used in other stages. In the future, we can employ a deep learning model in the meta-learning stage. In other words, reorganizing the output of the base learners to train a neural network may yield better results.
    \item Our method employs general architectures and a non-specific ensemble strategy, which means it is not limited to cervical cell classification and can be applied to other medical image classification tasks.
\end{itemize}

\section{Conclusion}
This paper proposes a Voting-Stacking ensemble of Inception networks for cervical cytology classification, mainly consisting of two stages. The first stage is data preprocessing, which includes resizing and data augmentation. The second stage is the Voting-Stacking ensemble, which selects three Inception family networks as base learners for the ensemble. The output is filtered and sent to train a meta-learner for the final cervical cell classification. We evaluate the proposed method on three benchmark datasets: SIPaKMeD, Herlev, and Mendeley, and achieve better performance in accuracy, recall, precision, and F1 score than the current SOTA methods. Moreover, we conduct experiments on super ensemble, which further improve the accuracy by training on the output of the meta-learner. This demonstrates that our method can effectively improve the accuracy of cervical pathology classification and has promising prospects for future applications in computer-aided diagnostic systems.

\bibliographystyle{unsrtnat}
\bibliography{references}  %%% Uncomment this line and comment out the ``thebibliography'' section below to use the external .bib file (using bibtex) .





% \end{thebibliography}

\end{document}